\documentclass[letterpaper, 10 pt, conference]{ieeeconf}  

\IEEEoverridecommandlockouts                              

\usepackage{graphics} 
\usepackage{epsfig} 
\usepackage{amsmath} 
\usepackage{amssymb}  
\usepackage{hyperref}
\usepackage{color}
\usepackage{multirow}
\usepackage[caption=false, font=footnotesize]{subfig}
\usepackage{threeparttable}
\newcommand{\bs}[1]{\boldsymbol{\mathrm{#1}}}

\title{\LARGE \bf
Learning Sequences of Manipulation Primitives for Robotic Assembly
}

\author{Nghia Vuong$^{1}$, Hung Pham$^{2}$, and Quang-Cuong
  Pham$^{1,2}$
  \thanks{$^{1}$Singapore Centre for 3D Printing (SC3DP), School of
    Mechanical and Aerospace Engineering, NTU, Singapore}
  \thanks{$^{2}$Eureka Robotics, Singapore}
}

\begin{document}

\maketitle
\thispagestyle{empty}
\pagestyle{empty}

\begin{abstract}
  This paper explores the idea that skillful assembly is best
  represented as dynamic sequences of Manipulation Primitives, and
  that such sequences can be automatically discovered by Reinforcement
  Learning. Manipulation Primitives, such as ``Move down until
  contact'', ``Slide along x while maintaining contact with the
  surface'', have enough complexity to keep the search tree shallow,
  yet are generic enough to generalize across a wide range of assembly
  tasks. Moreover, the additional ``semantics'' of the Manipulation
  Primitives make them more robust in sim2real and against
  model/environment variations and uncertainties, as compared to more
  elementary actions. Policies are learned in simulation, and then
  transferred onto a physical platform. Direct sim2real transfer
  (without retraining in real) achieves excellent success rates on
  challenging assembly tasks, such as round peg insertion with
  0.04\,mm clearance or square peg insertion with large hole
  position/orientation estimation errors.
\end{abstract}

\section{INTRODUCTION}

This paper explores the idea that skillful assembly is best
represented as dynamic sequences of Manipulation Primitives, and that
such sequences can be automatically discovered by Reinforcement Learning.

In recent years, increasingly complex assembly tasks have been
demonstrated on robot systems~\cite{suarez2018can}. However, in most
cases, the difficult assembly skills, such as tight pin insertion or
part mating, are still accomplished by hand-designed, hard-coded,
strategies (e.g. spiral search followed by force-controlled
insertion)~\cite{suarez2016framework}. Designing and fine-tuning such
strategies require considerable engineering expertise and time, thus
putting a brake on the deployment of intelligent robotic manipulation
in the factories and in the homes. This paper investigates how to
automatically discover \emph{in silico} assembly strategies that
robustly transfer to physical robots.

\subsection*{Representation of assembly skills as sequences of Manipulation
  Primitives}

The first, crucial, question is the representation of the assembly
skills: what is the set of atomic actions to be reasoned upon?
In~\cite{inoue2017deep}, the authors consider very simple atomic
actions such as pure force-controlled translations or pure
position-controlled rotations. This results in extremely long
sequences of atomic actions to achieve a given task, making the search
complexity overwhelming.

Consider how a robot would learn to play chess. One option is to learn
directly the sequences of robot commands to physically move the pieces
throughout the full game. Alternatively, it would be much more
efficient to learn the sequences of piece moves (e.g. 1. e4, 2. Nf3,
3. Bb5\dots), and then rely on grasp planning, inverse kinematics,
inverse dynamics, etc. to physically realize the moves.

\begin{figure}[!t]
  \centering
  \includegraphics[width=0.8\columnwidth]{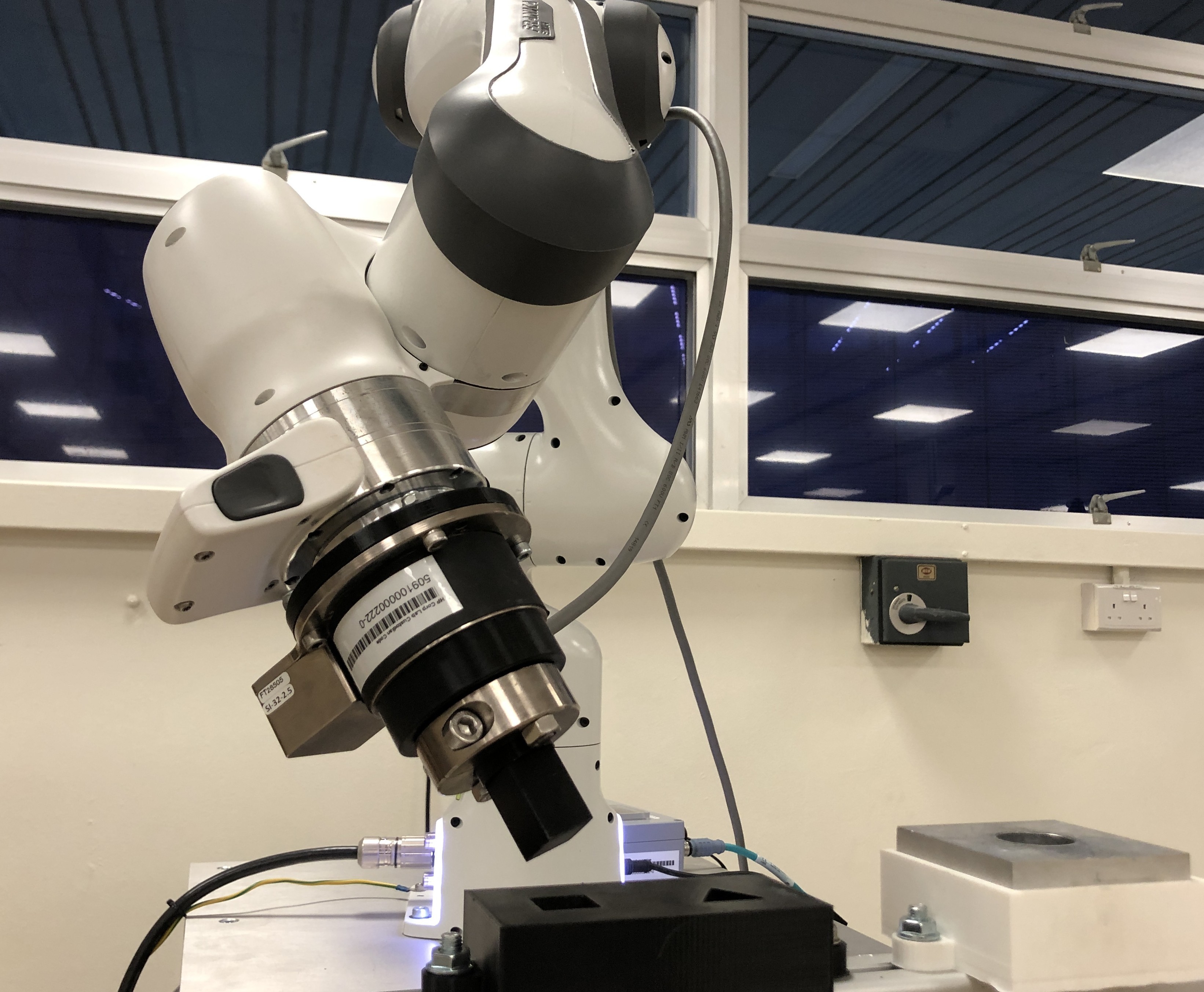}
  \caption{Robotic assembly setup. The video of the experiments is available at \href{https://youtu.be/P0NNjjQNOVo}{https://youtu.be/P0NNjjQNOVo}}
  \label{visual-id}
\end{figure}

Here, we propose, by analogy, to consider Manipulation Primitives
(MP)~\cite{johannsmeier2019framework} as the atomic
actions. Manipulation Primitives, such as ``Move down until contact'',
``Slide along x while maintaining contact with the surface'', have
enough complexity to keep the search tree shallow (typically a
sequence of 6 to 8 MPs is enough to achieve tight insertion), yet are
generic enough to generalize across a wide range of assembly tasks
(peg insertion with different peg shapes, large hole estimation
errors, random initial positions\dots) Another key advantage of MPs is
their additional \emph{semantics}, which make them robust in sim2real
and against model/environment variations and uncertainties: consider
how ``Move down until contact'' is inherently more robust than a
sequence of several short ``Move down'' actions.

\subsection*{Contribution: learning dynamic sequences of Manipulation
  Primitives}

In~\cite{johannsmeier2019framework}, the authors consider a set of MPs
with tunable parameters, the parameters being optimized through task
execution on the physical platform. However, the \emph{temporal
  sequence} of MPs to accomplish a given task is manually designed and
fixed, which re-raises the initial concern about expertise and time
required to address new tasks.

By contrast, we propose here to \emph{automatically discover dynamic
  sequences} of MPs by Reinforcement Learning (RL). Policies are
learned in simulation, and then transferred onto the physical
platform. We show that direct sim2real transfer (without retraining in
real) achieves 100\% and 95\% success rate on round peg insertion with
respectively 0.1\,mm and 0.04\,mm clearance, and despite 0.5\,mm and
0.5\,deg errors in hole position/orientation estimation. On the harder
task of square peg insertion with 1\,mm clearance, 1.5\,mm and
1.5\,deg errors in hole position/orientation estimation, direct
sim2real transfer still achieves 75\% success rate.

The rest of the paper is organized as follows. In Section
\ref{sec-review}, we review works that are related to our proposed
approach. In Section \ref{sec-controller-mp}, we formally define the
Manipulation Primitives. In Section \ref{sec-rl}, we introduce in
detail the proposed Reinforcement Learning formulation. In Section
\ref{sec-exp}, we present the experimental setup and quantitative
results. Finally, in Section \ref{sec-sum}, we discuss the advantages
and limitations of the presented approach, as well as some directions
for future work.

\section{RELATED WORK} \label{sec-review}

\textbf{Manipulation Primitives in robotic assembly.} Manipulation
primitive or skill primitive is a well-known concept in robot
manipulation. An advantage of programming with manipulation
primitives is to "move beyond the low-level representations of the
robot’s movements (classically joint-space or task-space and enable
generalizing robot capabilities in terms of elemental actions that can
be grouped together to complete any task"
\cite{suarez2016framework}. Recently, Johannsmeier et
al. \cite{johannsmeier2019framework} represents an assembly skill as a
directed graph whose nodes are MPs. They show that with the optimized
parameters, the graph efficiently performs several cylindrical
peg-in-hole tasks, even faster than human. However, the MPs are
designed manually, thus lack the ability to generalize to different
contexts. Furthermore, since the graph is generated offline, it could
not adapt to environmental uncertainties that might occur during
execution: one failure of any MPs might lead to the failure of the
whole execution. In our method, we do not assume a fixed sequence of
MPs; instead, the MP is generated at runtime.

\textbf{Deep RL in high precision robotics assembly.} Since
high-precision assembly is a challenging task, making reinforcement
learning tractable in this context requires careful design and
consideration. In the following discussion, we review various ways
reinforcement learning is made tractable in the context of
high-precision assembly.

One technique that has been successfully applied to high-precision
assembly task is smart action space discretization and problem
decomposition. In
\cite{inoue2017deep}, Inoue et al. decompose the task into a search
phase and an insertion phase, and designs different state space, action
space, and reward function for each phase. More specifically, several
meta-actions are designed to form a discrete action space. This
greatly speeds up training and achieves good performance at the same
time. However, one drawback of this method is flexibility, since a few
meta-actions limit the dynamic capabilities of the robot. We address
this problem by using a large set of such meta-actions. Although this
choice compromises the reduced training time, we adopt sim2real to
address this issue and argue that the use of MPs as meta-actions makes
the sim to real transfer more efficient. In the same vein, Hamaya et
al. \cite{hamaya2020learning} divide the peg-in-hole task into five
steps with different action spaces and state spaces. This
decomposition greatly speeds up training through dimensional reduction
of action and state spaces. The method, however, assumes a fixed
sequence of steps that might not generalize well to other
tasks.

In \cite{luo2019reinforcement}, Luo et
al. combines iterative Linear-Quadratic-Gaussian (iLQG)
\cite{todorov2005generalized} with an operational space force
controller to learn local control policies, then train a neural
network that generalizes this controller to adapt to environmental
variations, taking into account the force feedback. This method is
fast, being able to find a good control policy in just a few
interactions, but relies on iLQG to find the local controllers that
might not achieve a good performance on complex tasks, e.g. tight
insertion tasks, due to the imposed linear structure on the system
dynamic.

In \cite{schoettler2019deep}, Schoettler et al. study the use of image
observations and natural sparse rewards in several connector insertion
tasks. They also compare two techniques for incorporating prior
knowledge in RL: residual policy learning \cite{johannink2019residual}
and using demonstrations to guide the exploration of the subsequent RL
algorithm. Furthermore, they reduce the dimensions of the action
space: an action is a 3-dimensional position command. This implies
that the parts are aligned in the first place, which make the task
much easier. In fact, including orientation command results in
interesting strategies, as can be seen in our experimental results.

One line of research utilizes sim to real (sim2real) transfer to
reduce learning time on the real robot. Recent studies demonstrate
impressive sim2real results in the context of high-precision assembly
task. \cite{kaspar2020sim2real} applies system identification to align
the several simulation parameters (gravity, joint damping, etc.) with
the real robot dynamics. In \cite{schoettler2020meta}, a meta RL
algorithm called probabilistic embedding for actor-critic RL (PEARL)
\cite{rakelly2019efficient} is applied to learn the task structure for
a family of related tasks in simulation and adapt quickly to a real
task with few training data. Beltran et al. \cite{beltran2020variable}
use residual policy learning in combination with domain
randomization.

\section{MANIPULATION PRIMITIVES} \label{sec-controller-mp}

\subsection{Definition}
We follow \cite{johannsmeier2019framework} to define manipulation
primitives (MPs). An MP represents a desired motion of the robot
end-effector ($E$) in the task frame ($T$). More precisely, it
consists of:
\begin{itemize}
\item a desired velocity command $^{T}\bs{v}_{E}$ (in short $\bs{v}_{\mathrm{des}}$);
\item a desired force command $^{T}\bs{f}_{E}$ (in short $\bs{f}_{\mathrm{des}})$;
\item a stopping condition $\lambda$.
\end{itemize}

The desired velocity and force commands are defined as
\begin{align}
 \begin{split}
   \bs{v}_{\mathrm{des}}(t) &:= g_v(t, \bs{\Omega}_t;\bs{\theta}_v), \\
   \bs{f}_{\mathrm{des}}(t) &:= g_f(t, \bs{\Omega}_t;\bs{\theta}_f),
 \end{split}
 \label{equation:primitive}
\end{align}
where $g_v$ and $g_f$ are any functions parameterized respectively by
$\bs{\theta}_v$ and $\bs{\theta}_f$, and $\bs{\Omega}_t$ is the vector
of all sensor signals at time $t$. Finally, the stopping condition is
defined as
$\lambda : (t,\bs{\Omega}_t) \mapsto \{\verb+SUCCESS+, \verb+FAILURE+,
\verb+CONTINUE+\}$.
The next section instantiates our definition in the context of
peg-in-hole insertion tasks and clarifies the motivations.

\subsection{MPs for peg-in-hole insertion tasks} \label{subsec-mpexp}

For insertion tasks, we consider two families of MPs: free-space MPs
and in-contact MPs:
\begin{itemize}
\item Free-space MPs are to be executed when the robot is not in
  contact with the environment, i.e., when all external forces/torques
  are zero. MPs in this family are then associated with zero desired
  force/torque command.
\item In-contact MPs are to be executed when the robot is in contact
  with the environment, i.e. when some of the external force/torque
  components are non-zero. In addition to other objectives, MPs in
  this family have some of the components of their desired
  force/torque command to be non-zero in order to maintain the same
  contact state during the execution.
\end{itemize}

\begin{figure}[!t]
  \centering
  \subfloat[Free-space, translate until contact]{
    \label{fig-primitive-a}
    \includegraphics[width=0.4\columnwidth]{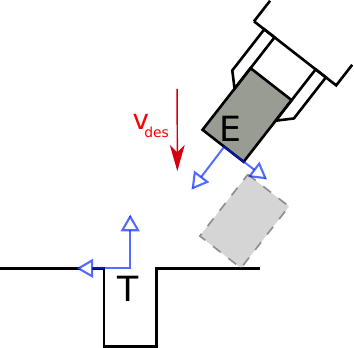}}
  \hspace{0.1\columnwidth}
  \subfloat[Free-space, translate a predefined distance]{
    \label{fig-primitive-b}
    \includegraphics[width=0.4\columnwidth]{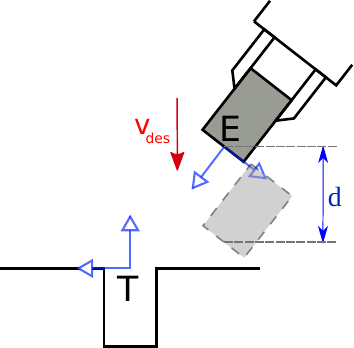}}\\
  \subfloat[In-contact, rotate until next contact]{
    \label{fig-primitive-c}
    \includegraphics[width=0.27\columnwidth]{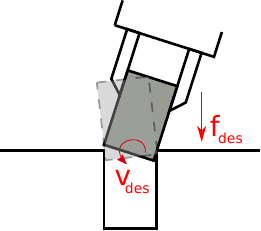}}
  \hspace{0.1\columnwidth}
  \subfloat[In-contact, insert]{
    \label{fig-primitive-d}
    \includegraphics[width=0.27\columnwidth]{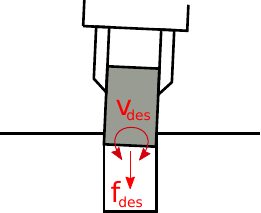}}
  \caption{Examples of Manipulation Primitives for insertion task. See
    text for details.}
  \label{fig-primitive}
\end{figure}

Each family of MPs are sub-divided into several types: (i) move until
(next) contact, (ii) move a predefined amount, (iii) insert. Figure
\ref{fig-primitive} illustrates some examples of MPs, which are
further detailed below. The set of all 91 MPs that are used in our
experiments is given in Table~\ref{tab-mp2}.

\textbf{Free-space, move until contact.} Translate the end-effector
along a direction, or rotate the end-effector about a direction, until
contact is detected. The example of (Fig.~\ref{fig-primitive-a})
translates the end-effector in the $-z$ direction with speed $v$ until
the measured force is larger than $f_\mathrm{thr}$ (\texttt{SUCCESS}),
or $t>2s$ (\texttt{FAILURE}), which is formally defined by
\begin{align}
  \begin{split}
    \bs{v}_{\mathrm{des}}(t) &= [0, 0, -v, 0, 0, 0]\\
    \bs{f}_{\mathrm{des}}(t) &= \bs{0}\\
    \lambda(t) &= \begin{cases} \texttt{SUCCESS} & \mbox{if } \bs{f}_{\mathrm{ext}}^T\bs{u}_v>f_\mathrm{thr},\\
      \texttt{FAILURE} & \mbox{if } t > 2, \\
      \texttt{CONTINUE} & \mbox{otherwise.}\end{cases}
  \end{split}
\end{align}
where $\bs{f}_{\mathrm{ext}}$ is the measured external force,
$\bs{u}_v = \bs{v}_{\mathrm{des}}/||\bs{v}_{\mathrm{des}}||$ is the moving direction.

\textbf{Free-space, move a predefined amount.} Translate the
end-effector along a direction over a predefined distance~$d$, or
rotate about a direction over a predefined angle $\alpha$. The example
of (Fig.~\ref{fig-primitive-b}) translates the end-effector in the
$-z$ direction with speed $v$, until the distance $d$ is reached
(\texttt{SUCCESS}), or a large contact force is detected
(\texttt{FAILURE}), which is formally defined by
\begin{align}
  \begin{split}
    \bs{v}_\mathrm{des}(t) &= [0, 0, -v, 0, 0, 0]\\
    \bs{f}_\mathrm{des}(t) &= \bs{0}\\
    \lambda(t) &= \begin{cases} \texttt{SUCCESS} & \mbox{if } \Delta\bs{p}^T\bs{u}_v>d\\
      \texttt{FAILURE} & \mbox{if } \bs{f}_s^T\bs{u}_v> f_\mathrm{thr}\\
      \texttt{CONTINUE} & \mbox{otherwise }\end{cases}
  \end{split}
\end{align}
where $\Delta\bs{p}$ is the distance between the current pose $\bs{p}$
and the start pose $\bs{p}_0$.

\textbf{In-contact, move until next contact.} Track a non-zero force
in a some directions, and translate the end-effector along a
direction, or rotate the end-effector about a direction until next
contact is detected. The example of Fig \ref{fig-primitive-c} control
a force $f_d$ in the $-z$ direction and rotate the peg around the $x$
direction with speed $v$, until the measured force is larger than
$f_\mathrm{thr}$ (\verb+SUCCESS+), or $t>2s$ (\verb+FAILURE+), which
is formally defined as
\begin{align}
  \begin{split}
    \bs{v}_{\mathrm{des}}(t) &= [0, 0, 0, 0, v, 0]\\
    \bs{f}_{\mathrm{des}}(t) &= [0, 0, -f_d, 0, 0, 0]\\
    \lambda(t) &= \begin{cases} \texttt{SUCCESS} & \mbox{if } \bs{f}_{\mathrm{ext}}^T\bs{u}_v>f_\mathrm{thr},\\
      \texttt{FAILURE} & \mbox{if } t > 2, \\
      \texttt{CONTINUE} & \mbox{otherwise.}\end{cases}
  \end{split}
\end{align}

\textbf{In-contact, insert.} Track a non-zero force in the direction
of insertion and regulate the forces and torques to zero in all the
other directions.  The example
of (Fig.~\ref{fig-primitive-d}) performs the insertion in the $z$
direction, as formally defined by
\begin{align}
  \begin{split}
    \bs{v}_{\mathrm{des}}(t) &= -K_d\bs{f}_{\mathrm{ext}}\\
    \bs{f}_{\mathrm{des}}(t) &= [0, 0, f_d, 0, 0, 0]\\
    \lambda(t) &= \begin{cases} \texttt{SUCCESS} & \mbox{if } d(\bs{p}, \bs{p}_t) < \epsilon \\
      \texttt{FAILURE} & \mbox{if } t > 2\\
      \texttt{CONTINUE} & \mbox{otherwise }\end{cases}
  \end{split}
\end{align}
where $K_d$ is a compliant $6\times6$ diagonal matrix, $d(.)$ is a
metric measuring distance between two poses, $\bs{p}_g$ is the goal
pose. For simplicity, we consider a diagonal compliant matrix of the
form $K_d = \left[ \begin{matrix}
    k_{dt}\mathbb{I}^{3\times 3} & \bs{0}\\
    \bs{0} & k_{dr}\mathbb{I}^{3\times 3}
  \end{matrix} \right]$.

\begin{table*}[!ht]
  \vspace*{5pt}
  \caption[caption]{The set of 91 Manipulation Primitives used in our experiments.}
  \label{tab-mp2}
  \begin{center}
    \begin{threeparttable}
      \begin{tabular}{|c|c|c|c|c|c|}
        \hline
        Family & Type & Axis & Parameters & Values & N\\
        \hline
        \multirow{8}{*}{Free space} & \multirow{2}{*}{Translate until contact (T$^\textrm{c}$)} & \multirow{2}{*}{$-z$} & $v$ & 10\,mm/s & \multirow{2}{*}{1}  \\
               &    &    & $f_\mathrm{thr}$ & 8\,N  &   \\
        \cline{2-6}
               & \multirow{3}{*}{Translate (T)} & \multirow{3}{*}{$\pm x,\pm
                                                  y,\pm z$} &
                                                              $v$ & 10\,mm/s & \multirow{3}{*}{12} \\
               & & & $f_\mathrm{thr}$ & 15\,N & \\
               & & & $d$ & 2 or 4\,mm & \\
        \cline{2-6}
               & \multirow{3}{*}{Rotate (R)} & \multirow{3}{*}{$\pm x,\pm
                                               y,\pm z$} &
                                                           $v$
                                          & 9\,deg/s & \multirow{3}{*}{12} \\
               & & & $f_\mathrm{thr}$ & 1\,Nm & \\
               & & & $d$ & 2 or 4\,deg & \\
        \hline
        \multirow{16}{*}{In contact} & \multirow{3}{*}{Translate
                                       until next
                                       contact (T$^\textrm{c}$)}& \multirow{3}{*}{$\pm x, \pm y$} &  $v$ & 4 or 7.5\,mm/s & \multirow{3}{*}{16}\\

                                       & & & $f_\mathrm{thr}$ & 8 or 15\,N & \\
                                       & & & $f_d$ & -3\,N & \\
        \cline{2-6}
         & \multirow{3}{*}{Rotate
                           until next
                           contact (R$^\textrm{c}$)}& \multirow{3}{*}{$\pm x, \pm y, \pm z$} &  $v$ & 4 or 7\,deg/s & \multirow{3}{*}{24}\\

               & & &  $f_\mathrm{thr}$ & $0.1$ or $0.5$\,Nm & \\

               & & &  $f_d$ & $-3$\,N & \\
        \cline{2-6}
               & \multirow{4}{*}{Translate (T)} & \multirow{4}{*}{$\pm x, \pm y$} & $v$ & 10\,mm/s  & \multirow{4}{*}{8} \\

               & & & $f_\mathrm{thr}$ & 15\,N & \\
               & & & $d$ & 2 or 4\,mm & \\
               & & & $f_d$ & $-3$\,N & \\
        \cline{2-6}

              & \multirow{4}{*}{Rotate (R)} & \multirow{4}{*}{$\pm x, \pm y, \pm z$} & $v$ & 4.6\,deg/s  & \multirow{4}{*}{12} \\

              & & & $f_\mathrm{thr}$ & 1\,Nm & \\
              & & & $d$ & 2 or 4\,deg & \\
              & & & $f_d$ & $-3\,$N & \\
        \cline{2-6}
               & \multirow{4}{*}{Insert (I)} & \multirow{4}{*}{$-z$} & $\epsilon$ & 2mm & \multirow{4}{*}{4} \\
               &  &  & $k_{dt}$ & 0.01 &  \\
               &  &  & $k_{dr}$ & 0.05 or 0.1 &  \\
               &  &  & $f_d$ & $-5$ or $-12$\,N &  \\
        \hline
      \end{tabular}
    \end{threeparttable}
  \end{center}

  \vspace*{-10pt}
\end{table*}

\section{LEARNING DYNAMIC SEQUENCES OF MANIPULATION PRIMITIVES BY RL}
\label{sec-rl}

The MPs sequencing problem is to find
sequences in a finite set $\mathbb{M}$ of MPs that successfully perform a
particular task. More specifically, we are interested in the family of
insertion tasks, which is characterized by a goal configuration.

\subsection{Overview of Reinforcement Learning}
We consider here the discounted episodic RL problem. In this setting,
the problem is described as a Markov Decision Process (MDP)
\cite{sutton2018reinforcement}. At each time step $t$, the agent
observes current state $\bs{s}_t \in \mathbb{S}$,
executes an action $\bs{a}_t \in \mathbb{A}$, and receives an immediate
reward $r_t$. The environment evolves through the state transition
probability $p(\bs{s}_{t+1} | \bs{s}_t, \bs{a}_t)$. The goal in RL is to learn a
policy $\bs{a}_t=\pi (\bs{s}_t)$ that maximizes the expected discounted return
$R = \sum_{t=1}^{T} \gamma^t r_t$, where $\gamma$ is the discount
factor that tends to emphasize the importance of most recent rewards.
\subsection{Our RL setting}

\textbf{State and action.} An action is the index of an MP to be
executed. This results in a discrete action space since the MP set $\mathbb{M}$
contains a finite number of MPs. This action set consists of
two subspaces: $\mathbb{A}_\mathrm{free}$ containing all free-space MPs and $\mathbb{A}_\mathrm{con}$ containing all in-contact MPs. We define the state vector
$\bs{s}_t = [\bs{p}(t), \bs{f}_{ext}(t)]$. where $\bs{p}$
is the relative pose between the two involved parts, computed
from the joint position, the goal pose, and forward kinematics;
$\bs{f}_{ext}$ is the external force and torque acting on the
end-effector. All of the measurement and calculation at time $t$
are performed after the execution of the MP at
time $t-1$, i.e. the stop condition $\lambda$ returns \verb+SUCCESS+
or \verb+FAILURE+. The set of feasible actions at each state $\bs{s}_t$ is
either $\mathbb{A}_\mathrm{free}$ if $\bs{f}_\mathrm{ext}=\bs{0}$, or $\mathbb{A}_\mathrm{con}$ if $\bs{f}_\mathrm{ext}\neq\bs{0}$.

\textbf{Algorithm and policy parameterization.} We use Proximal Policy
Optimization (PPO) \cite{schulman2017proximal}, an on-policy and
model-free RL algorithm. Due to its noisy nature, we do not use force
measurement as the input to the policy and value functions. The value
function is a Multilayer Perceptron (MLP). For the policy, we use two
separate neural networks for each of the action subspace.

\textbf{Reward function.} We define the reward function as
\setlength{\arraycolsep}{0.0em}
\begin{eqnarray}
  r(\bs{o}_t, \bs{o}_{t+1}, \bs{a}_t)&{}:={}&c_1\left(e^{\frac{-||\bs{p}_{t+1} - \bs{p}_{goal}||_2^2}{k_1}} -1\right)\nonumber\\
  &&{-}\:c_2t(\bs{a}_t) + c_3s(\bs{a}_t)
  \label{eq-rew}
\end{eqnarray}
\setlength{\arraycolsep}{5pt}

The first term rewards for moving closer to the goal, the second term
is the execution time of the MP, to bias the algorithm to find
solution with short execution time. The third term $s(\bs{a}_t)=0$ if
a \verb+SUCCESS+ status is returned, $s(\bs{a}_t)=-1$ if a \verb+FAILURE+
status is
returned. 

\section{EXPERIMENTS} \label{sec-exp}

\subsection{System description}

\textbf{Task description.} We design multiple peg-in-hole tasks with three
types of peg profiles: round shape, square shape, and triangular
shape. The properties of the pegs are shown in Table \ref{tab-dim} We
make the following assumptions regarding the insertion tasks: (1) The
peg is firmly mounted to the robot end-effector (2) The hole is fixed
in the environment (3) An estimation of the relative position and
orientation between the peg and hole can be obtained, either by a
vision system or by teaching.

\begin{table}[!t]
\vspace*{5pt}
\caption[caption]{Dimensions and material of pegs and holes}
\label{tab-dim}
\begin{center}
\begin{tabular}{|c|c|c|c|c|}
\hline
Profile & Hole size & Insertion depth & Material & Clearance \\
\hline
Round & 30\,mm & 20\,mm & Aluminum & 0.1\,mm \\
\hline
Round & 30\,mm & 20\,mm & Aluminum & 0.04\,mm \\
\hline
Square & 20.5\,mm & 20\,mm & Plastic & 1\,mm \\
\hline
Triangle & 25.1\,mm & 20\,mm & Plastic & 1\,mm \\
\hline
\end{tabular}
\end{center}
\end{table}

\begin{figure}[!t]
  \vspace*{5pt}
  \centering
  \subfloat[]{\includegraphics[scale=1.0]{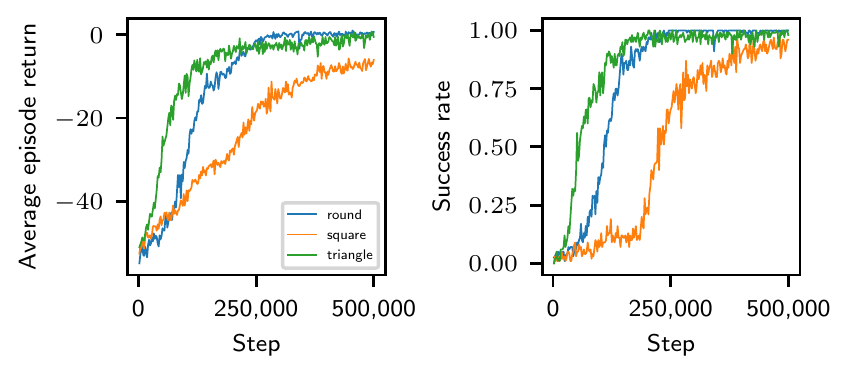}}\\
  \subfloat[]{\includegraphics[scale=1.0]{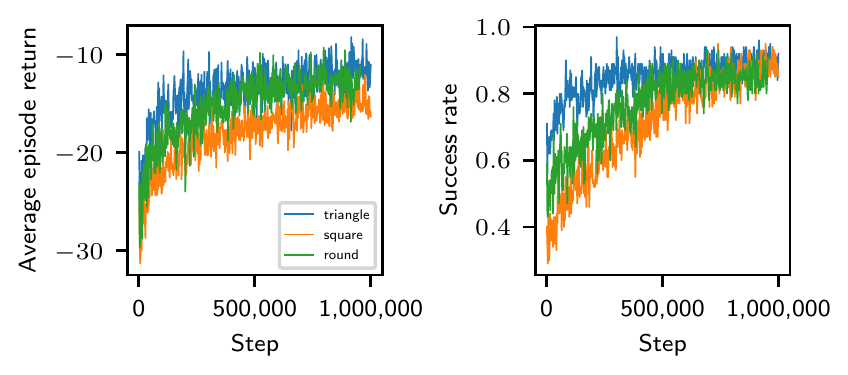}}
  \caption{Training curve (average episode reward and
    success rate) for (a) Training Condition TC1 and (b) Training
    Condition TC2.}
  \label{fig-training}
\end{figure}

\textbf{Robot system setup.} A 7-DOF Franka Emika Panda cobot is
used in our experiment. We use the Mujoco physics engine
 \cite{todorov2012mujoco} and adapt
an open-source Panda robot model \footnote{available online at
  \href{https://github.com/vikashplus/franka_sim}{franka\_sim}}. The
controller is simulated with a control frequency of 500Hz. In real
world, we additionally use a Gamma IP60 force torque sensor to measure
the external force acting on the end-effector. The measurement from FT
sensor is needed to implement insert primitive, as the external force
estimation in \verb+libfranka+ is not precise enough to perform this
primitive. We teach the position of the hole for real world
experiments

\textbf{Implementation details.} We use \verb+gym+ \cite{brockman2016openai}
 to design the
environment and \verb+rlpyt+ \cite{stooke2019rlpyt}, a RL framework
based on PyTorch for the implementation of PPO algorithm.

\subsection{Training in simulation}

We design the environments in Mujoco for
three peg-in-hole tasks with different peg and hole shapes: round,
square and triangle. The default starting position of the peg is above
the hole $10$\,mm. To improve the robustness and generalization capability of
the trained policy, at the start of each episode, we add a
displacement $\Delta\bs{p}_\mathrm{init}$ to the starting position and
hole estimation error $\Delta\bs{p}_\mathrm{hole}$ to the true hole
pose. For each task, we train on two different Training Conditions:

\begin{description}[\labelindent=0pt]
\item[(TC1)] $\Delta\bs{p}_\mathrm{init}$ is uniformly sampled in $(-1, 1)$\,mm
  for position and in $(-1, 1)$\,deg for orientation,
  $\Delta\bs{p}_\mathrm{hole} = \bs{0}$;
\item[(TC2)] $\Delta\bs{p}_\mathrm{init}$ is uniformly sampled in $(-2, 2)$\,mm
  for position and $(-2, 2)$\,deg for orientation,
  $\Delta\bs{p}_\mathrm{hole}$ is uniformly sampled in $(-1, 1)$\,mm for
  position and in $(-1, 1)$\,deg for orientation,
\end{description}

The weights of the policy trained for (TC1) is used to initialize the
policy trained with (TC2). We do not train directly on (TC2) due to
its difficulty. The success rate and average reward during training
are reported in Fig.~\ref{fig-training}.

\begin{figure}[!t]
  \vspace*{-5pt}
  \centering
  \includegraphics[]{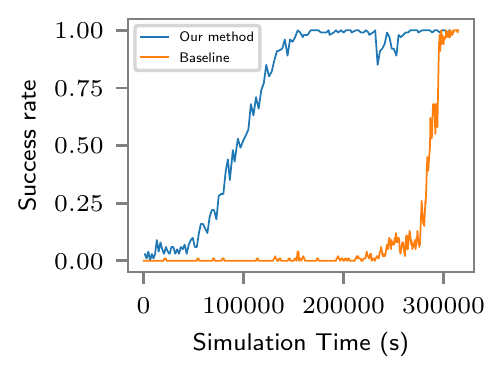}
  \caption{Comparison of learning performance between proposed method and the baseline.}
  \label{fig-baseline}
\end{figure}

We also compare our method with a baseline method, which learns in a
continuous action space. Concretely, the RL policy of the baseline
mathod takes as input the relative pose between the mating parts and
outputs the desired end-effector displacement. The displacement is
controlled by an impedance controller. The reward at each time step is
the sum of three components: (1) negative reward based on the distance
between the peg and hole; (2) negative reward if contact force is $>
30N$; (3) termination reward if the trial is successful. We train the
baseline method for the round peg-in-hole task under (TC1) and
illustrates the results in Fig.~\ref{fig-baseline}. As can be seen from the
figure, the baseline methods is significantly slower than our proposed
method. This suggests that using MPs improves the exploration at the
initial stage of learning, thanks to the more shallow search
tree. This advantage is also mentioned in the \emph{option framework}
\cite{barto2003recent}, which is related to our approach.

\subsection{Sim2real policy transfer on physical robot}

\begin{figure}[!t]
  \vspace*{-10pt}
  \centering
  \subfloat[Round, 0.1\,mm clearance]{\includegraphics[width=0.5\columnwidth]{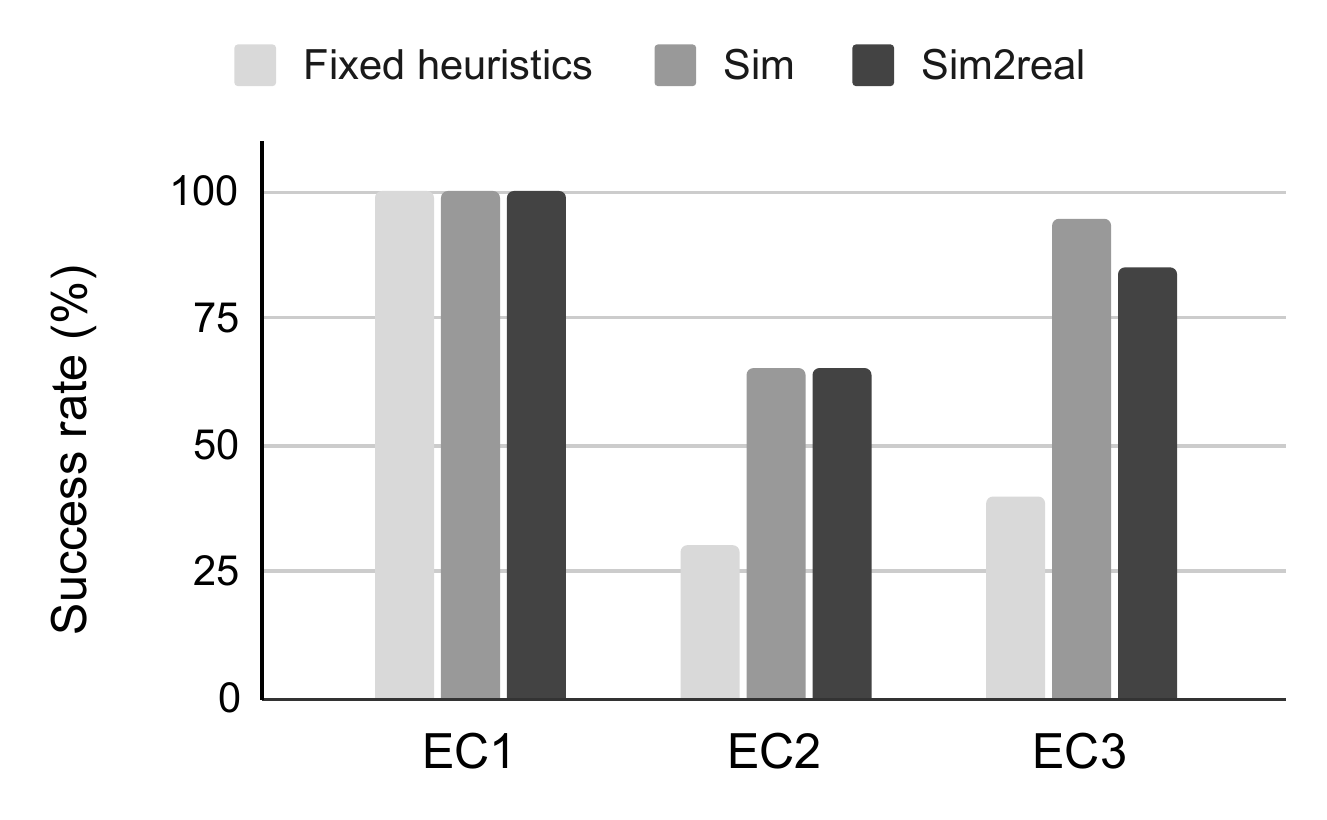}}
  \hfill
  \subfloat[Round, 0.04\,mm clearance]{\includegraphics[width=0.5\columnwidth]{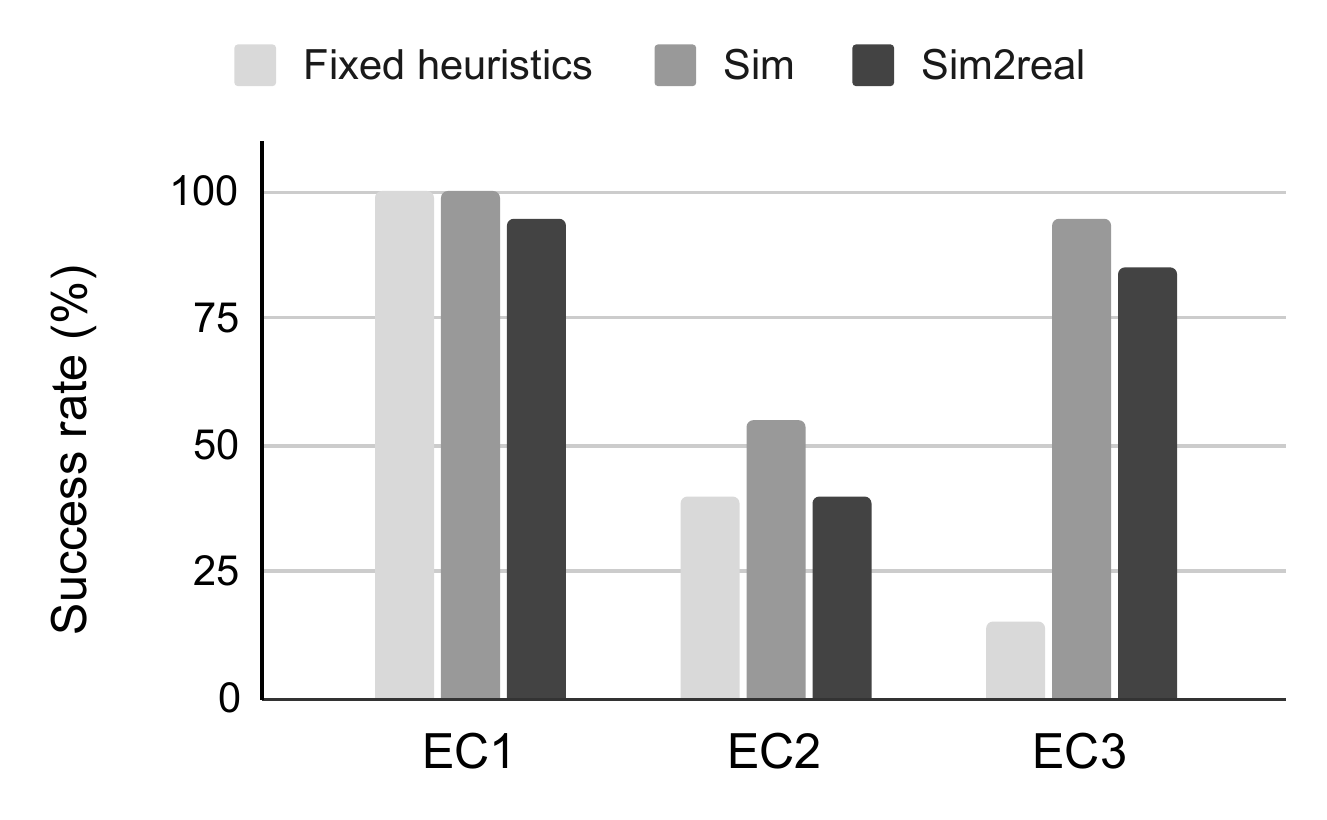}}\\

  \subfloat[Square]{\includegraphics[width=0.5\columnwidth]{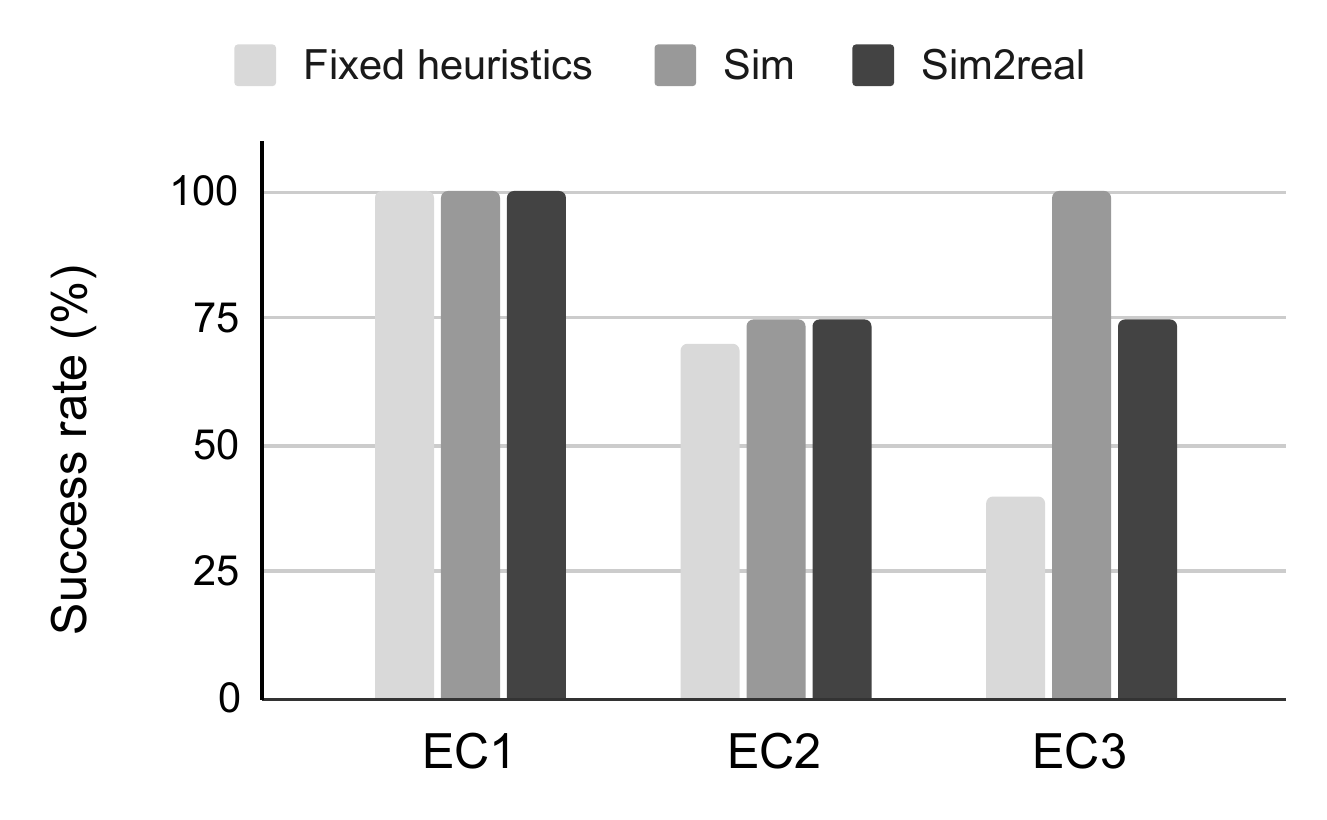}}\\

  \caption[caption]{Evaluation on the round and square peg-in-hole
    task on three Evaluation Conditions.}
  \label{fig-compare}
\end{figure}

We evaluate the learned policies directly on the real robot without
any further fine-tuning. The results are reported in
Fig~\ref{fig-compare}. We consider three Evaluation Conditions:

\begin{description}[\labelindent=0pt]
\item[(EC1)] Nominal performance: $\Delta\bs{p}_\mathrm{init}=\bs{0}$, $\Delta\bs{p}_\mathrm{hole}$ is sampled in $(-0.5, 0.5)$\,mm for position and in $(-0.5, 0.5)$\,deg for orientation;
\item[(EC2)] Generalizability: $\Delta\bs{p}_\mathrm{init}=\bs{0}$, $\Delta\bs{p}_\mathrm{hole}$ is sampled in $(-1.5, 1.5)$\,mm for position and in $(-1.5, 1.5)$\,deg for orientation;
\item[(EC3)] Robustness: $\Delta\bs{p}_\mathrm{init}$ is sampled in $(-1, 1)$\,mm
  for position and $(-1, 1)$\,deg for orientation,
  $\Delta\bs{p}_\mathrm{hole}$ is sampled in $(-0.5, 0.5)$\,mm for
  position and in $(-0.5, 0.5)$\,deg for orientation
\end{description}

Different from the training phase, the hole estimation error
and the initial pose displacement are sampled
on the \textit{boundary} of the box around the norminal values.

\begin{figure}[!t]
  \captionsetup[subfigure]{labelformat=empty}
  \vspace*{5pt}
  \centering
  \includegraphics[scale=0.4]{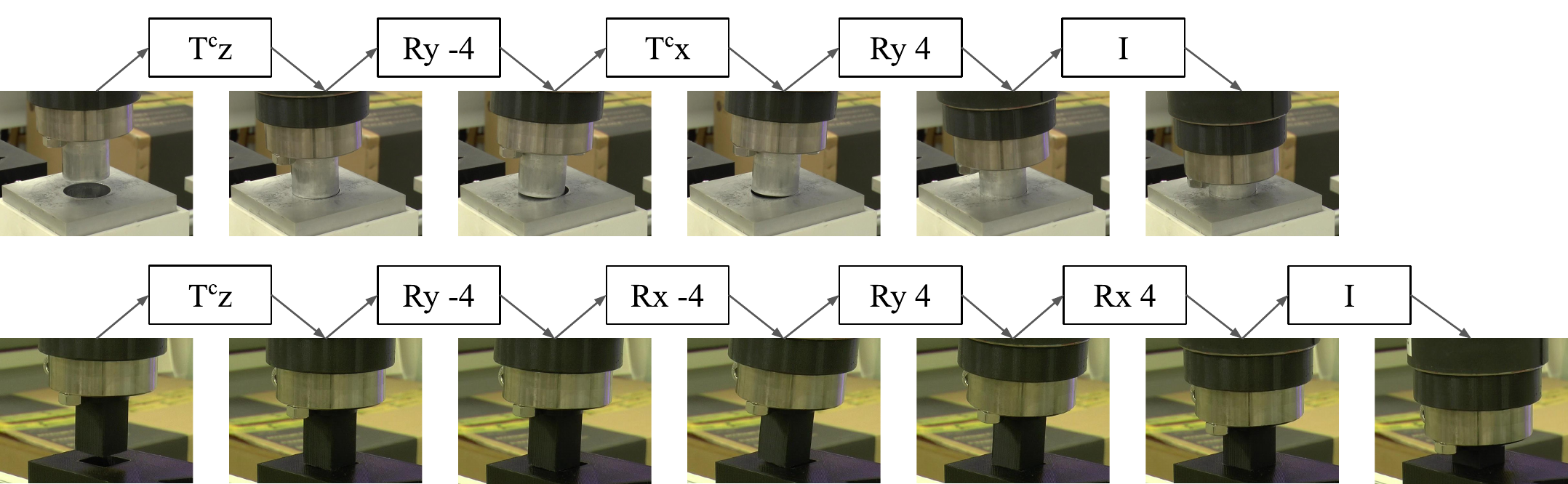}
  \caption{Snapshots of four runs on the round and square peg-in-hole
    insertion tasks. ``Ry 8'' means rotation of 8\,deg around y,
    which is the concatenation of two MPs that rotate 4\,deg each.
    Note the different sequences of MPs for the same task, which
    illustrates the \emph{dynamic} character of the learned policies. See the full video of these sequences at \href{https://youtu.be/P0NNjjQNOVo}{https://youtu.be/P0NNjjQNOVo}}
  \label{fig-snapshot}
 \end{figure}

We next compare the transfer result with a manually-defined sequence
of MPs and report the result in Fig~\ref{fig-compare}. This sequence
is tuned for the round peg-in-hole task with an estimation error of
$1$\,mm in translation and $1\,$deg in orientation. More specifically,
the sequence is (1) rotate about $-y$ $5\,$deg; (2) translate along
$-z$ until next contact; (3) translate along $x$ until next contact;
(4) rotate about $y$ until next contact; (5) insert. One can see that
the manually-defined solution does not generalize well: for both round
and square tasks, the success rates for Evaluation Condition EC2 are
significantly lower than our proposed method.

\begin{figure}[!t]
  \vspace*{-10pt}
  \centering
  \includegraphics[width=\columnwidth]{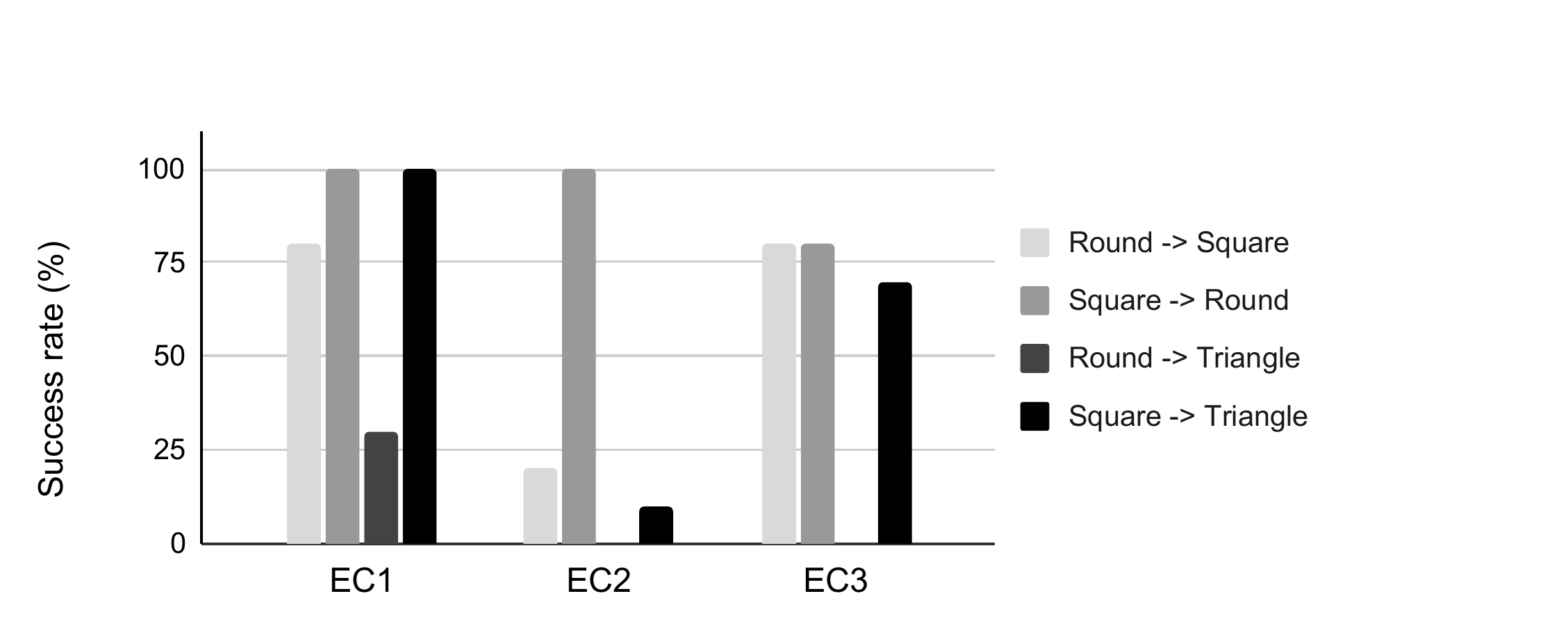}
  \caption{Results for the policy transfer experiments. $A \rightarrow
    B$: the policy trained for shape $A$ is evaluated on shape $B$.}
  \label{fig-transfer}
\end{figure}

We also run the trained policy on tasks with different shapes from the
one the policy was trained for, on $N=10$ trials. The results are
shown in Fig~\ref{fig-transfer}. The result demonstrates the
generalization capability of the trained policy across different
geometries of the parts. For instance, the policy trained for square
peg-in-hole confidently solves the round peg-in-hole task, even with
large estimation error.

\subsection{Dynamic character of the learned policies}

We investigate next the emergent behaviors exhibited by the trained
policies. All strategies tend to find a correct pose, such that the
insert MP could completes the task afterward. To achieve such pose,
the most commonly used MP is of rotation type (see Fig.~\ref{fig-snapshot} and video at \href{https://youtu.be/P0NNjjQNOVo}{https://youtu.be/P0NNjjQNOVo}). Rotating motion induces a tilted peg posture.
This posture effectively broadens the state spaces in which parts of
the peg are inside the hole. Interestingly, for the square peg-in-hole
task, the policy follows this strategy by rotating the peg in both $x$
and $y$ directions. After reaching such states, a "translate until
contact" often comes next to achieves the locally "optimal" position,
where the peg is at the lowest position (refer to two top rows of
Fig.~\ref{fig-snapshot}). After that, a rotating motion is regulated to
cancel the one in previous steps, before the insertion takes place.

\section{CONCLUSIONS} \label{sec-sum}

In this paper, we have explored the idea that skillful assembly is
best represented as dynamic sequences of Manipulation Primitives, and
that such sequences can be automatically discovered by Reinforcement
Learning. To illustrate this idea, we designed a set of MPs for
peg-in-hole insertion tasks, and proposed a sim2real approach:
policies are learned in simulation, and then transferred onto the
physical platform, without re-training in real. The experimental
results showed that policies learned purely in simulation were able to
consistently solve tight-clearance round peg insertion tasks, and
square peg insertion tasks with large estimation errors.

In future work, we shall evaluate our approach on more complex tasks,
such as a tight-clearance polygonal peg-in-hole insertion, multiple
pin insertion, connector insertion, gear assembly, DIMM memory
assembly, etc.

Another potential direction is to integrate tactile and visual
information, either to select the correct next MP, or to design a
visual-based MP, such as "visual servoing".

In tasks that have complex dynamics, where instability is a key
consideration, the choice of the low-level control law in each MPs
define the upper limit of the overall system performance. Hence,
incorporating advanced robust control laws~\cite{pham2020convex} is a
promising direction.




%

\section*{ACKNOWLEDGMENT}
This research was supported by the National Research Foundation, Prime
Ministrer’s Office, Singapore under its Medium Sized Centre funding
scheme, Singapore Centre for 3D Printing, CES\_SDC Pte Ltd, and Chip
Eng Seng Corporation Ltd.


%

\end{document}